# Have a Break from Making Decisions, Have a MARS: The Multi-valued Action Reasoning System

Cosmin Badea[0000-0002-9808-2475]

Imperial College London, London, SW7 2AZ, UK
c.badea@imperial.ac.uk

**Abstract.** The Multi-valued Action Reasoning System (MARS) is an automated value-based ethical decision-making model for artificial agents (AI). Given a set of available actions and an underlying moral paradigm, by employing MARS one can identify the ethically preferred action. It can be used to implement and model different ethical theories, different moral paradigms, as well as combinations of such, in the context of automated practical reasoning and normative decision analysis. It can also be used to model moral dilemmas and discover the moral paradigms that result in the desired outcomes therein. In this paper we give a condensed description of MARS, explain its uses, and comparatively place it in the existing literature.

**Keywords:** Logics for AI, Multi-criteria Decision-making, AI Ethics, Intelligent Decision Support Systems, Intelligent Agents, Expert and Knowledge-based Systems.

## 1 Background

What is the right thing to do in any given situation? Even though all of us face some instance of this question often, as a society we have not yet reached a consensus about morality, let alone about a formalization thereof (Sayre-McCord 2014). These days, however, with the increasing prevalence and sophistication of artificial agents, especially with regards to autonomous systems, finding an automated approach to moral reasoning is getting more urgent than ever.

Artificially intelligent (AI) agents in autonomous systems are by their very nature designed to react differently in different situations. This can be their main strength and at the same time a potential danger, as the designers of these systems cannot reasonably predict all possible situations the agent might be faced with. As they could be infinitely many, it seems impossible to define adequate responses to all of them *a priori*. How, then, can we ensure that an artificial agent's actions are in line with the moral values of the society which it is embedded in, to obtain moral or safe agents?

This is the main challenge in machine ethics (Anderson & Anderson 2011, Wallach 2016).

As part of answering this, we address the question: **Given the relevant circumstances, how should an AI agent automatically choose between different actions based on their underlying moral significance?**
It has been argued that we should build moral agents with inbuilt moral paradigms, allowing them to do explicit moral reasoning, and our work aims to go down this route and contribute to the creation of explicit moral agents (Wallach & Allen 2008). However, we show that the results can also be used to build implicit moral agents, whereby the moral dimension is brought by the rules of behaviour built into the agent and previous moral reasoning performed by the human designers of such systems, as opposed to moral reasoning by the agent itself.

Autonomous agents are already being used in applications such as self-driving cars, drones, trading applications and manufacturing (Bekey et al. 2008). Some researchers are working on extending the systems' abilities at a fundamental level and on exploring their potential in new fields such as search-and-rescue or patient and health care (Andrade et al. 2014, Hindocha & Badea 2022).

What might a scenario look like in which an AI agent would be required to make an ethical decision? A well-known example is the Trolley Problem (Foot 1967). This has been widely discussed in the literature in different areas such as philosophy (Foot 1967) and moral psychology (Greene 2001) and has recently attracted a lot of attention because of its relevance to autonomous vehicles (Bonnefon 2016).

Our research builds on previous work in moral philosophy and practical reasoning, machine ethics and logic of argumentation. Below we summarize related work from those fields.

## 1.1 Philosophical Underpinnings

Several different moral theories exist, but we have focused on the three that we believe are most relevant and important for this type of automated decision analysis.

The first theory is *virtue ethics* which, under the Aristotelian formulation, concerns itself with an individual's moral character as central to moral reasoning, and focuses on *arete* ("excellence", "virtue") (Aristotle 2009). Therein, moral virtue is seen as a disposition to behave in the right manner, and specific virtues are to be found in the mean between deficiency and excess. *Phronesis*, or practical wisdom, is achieved through habituation and practice and guided, ideally, by how the *phronimos* (the wise one) would act.

The second theory is *consequentialism*, which relies on the view that the rightness of actions depends on their consequences, and the formulation of it called *utilitarianism* (Sinnott-Armstrong 2015). The maximization of utility is core to this theory, and its calculation relies on considerations such as pleasure and the absence of pain.

The third theory is *deontology*, whereby it is held that one must act with the right intention and in accordance with duty, the moral norms of which can be obtained and reasoned through using one's autonomous will and rationality (Alexander & Moore 2016). Kantian deontology uses the Categorical Imperative as a tool in formulating perfect duties and imperfect duties.

### 1.2 Machine Ethics

Within the field of machine ethics, two different approaches to engineering moral decision-making have emerged: the top-down encoding of ethical theories and the bottom-up design of systems which aim at a specified goal (Wallach et al. 2008, Charisi et al. 2017). It has been argued that a combination of the top-down and bottom-up approaches will be required to successfully design artificial moral agents (Charisi et al. 2017, Allen, Smit, Wallach 2005).

**Bottom-up Approaches.** GenEth (Anderson & Anderson 2014), a general ethical dilemma analyzer, learns ethical principles from examples. There is more work by Anderson & Anderson on using a bottom-up approach, such as *W.D.* (Anderson, Anderson, Armen 2005), *EthEl* (Anderson & Anderson 2008) and *MedEthEx* (Guarini 2011). Other approaches, employing case-based reasoning, are *Truth-Teller* (McLaren & Ash-ley 1995) and *Sirocco* (McLaren 2003). Also, Guarini (Guarini 2011) trains a Recurrent Neural Network to learn the moral permissibility of actions involving allowing to die and killing.

**Top-down Approaches.** Dennis et al. (Dennis et al. 2016) explore ways of making automated ethical reasoning formally verifiable. In their book, Pereira et al. (Pereira et al. 2016) investigate both the individual realm of ethics, i.e., how humans make moral decisions on a personal level, and the collective realm of ethics.

**Value-based Argumentation.** Bench-Capon (Bench-Capon 2002) introduces values to practical reasoning in argumentation frameworks as commonly applied in a legal and moral context.

## 2 Multi-valued Action Reasoning System

In the present work, we focus on the decision-making part of the problem as well as on the formalization of moral paradigms. As such, we assume a system with well-defined actions to choose between and with some relevant considerations (which we call *values*).

For brevity and due to the format and space requirements, we shall omit some figures and formulas here. The publication of MARS in full detail is forthcoming.

Firstly, we ask: what would be important aspects of such an ethical decision-making framework? Most importantly, correctness should be established. That is, in any given situation the system should make the decision which we want it to. To test this, we use examples that have already been discussed in the literature, so that we have a way of comparing our results and verifying that they match what is expected. Furthermore, we also want an ethical reasoning system that is transparent, in the sense of our being able to track its reasoning process. The explicit structured representation of moral paradigms presented herein enables us to trace a decision to the underlying factors, namely the values, the moral paradigms, and their interactions. Lastly, we want the framework to be as flexible as possible, allowing us to model many different moral paradigms. The *Multi-valued Action Reasoning System (MARS)* which we present herein aims to address these aspects of automated moral reasoning.

## 2.1 Practical Motivating Example

We will illustrate the functioning of the system using the example of *Hal the diabetic*, as formulated in (Bench-Capon 2002):

> Hal, who suffers from diabetes and urgently needs insulin to survive, has lost his supply through no fault of his own. Carla, who also suffers from diabetes, does have insulin and Hal can take it from her house, without her permission or knowledge. Knowing that Carla has diabetes too, would he be justified in taking the insulin?

To model the dilemma of Hal the diabetic, we first need a way of representing Hal's options.

An **action** $a_i$ is a process that can be selected by an agent in the context of a decision. Semantically, choosing an action means electing to undertake the corresponding act in the situation represented by the decision. We hold the assumption that there is a finite number of mutually exclusive actions under consideration, and we denote the set of actions as $A = \{a_1, a_2, \ldots a_n\}$. Mutual exclusivity here refers only to the scope of the decision problem under consideration. That is, only one of the actions can be performed by the agent, and its occurrence implies the non-occurrence of all other actions.

In Hal's case, we follow the source (Bench-Capon 2002) and set the available actions to $A = \{$"take insulin", "don't take insulin"$\}$.

## 2.2 Values

Given these actions, we need a way of assessing them through the lens of an agent's moral paradigm. Values have been suggested as a suitable basis for ethical reasoning (Anderson & Anderson 2011). We subscribe to the argument for using a value-based approach (Bench-Capon 2002) and believe that using them allows us to evaluate an action through a given moral theory by making a concrete connection between the two. An action can be analyzed in terms of relevant values, and a moral theory can also be represented in terms of values, given a suitable framework, such as the one which we present in our work.

Our concept of values aims to be as general as possible, so that we can model different philosophical approaches. That means the chosen values could in practice be of myriad sorts: reasons for or against performing actions, consequences, intentions, duties, principles or rules, virtues, states of affairs to obtain, possible worlds or features of such etc. Not all the above need to be used in all situations. Thus, for example, a purely consequentialist approach can be modelled by only looking at values as consequences.

Essentially, in MARS a **value** $v_i$ is a concept which is considered to be of (moral) significance in the given scenario. Values are the first input for MARS. The finite set of values we denote with $V$.

In our *Hal the Diabetic* example, similarly to the source (Bench-Capon 2002), we might identify "Hal's life", "Carla's life" and "Property rights", or "Property", as the values. These are relevant values because taking the insulin would save Hal's life, but endanger Carla's, while not taking it could be seen as respecting her property rights. So our $V$ = {"Hal's life", "Carla's life", "Property"}.

To talk about the impact that actions have on values, we use the concepts of *promotion* and *demotion*. Such concepts have been proposed before, for example in (Anderson & Anderson 2011, Anderson & Anderson 2014, Feldman 1997).

We say that action $a$ **promotes** value $v_1$ to mean that performing this action is in the spirit of the value. Conversely, we say that action $a$ **demotes** value $v_2$ to mean that performing this action is against the spirit of the value. We denote the two sets of values that are promoted and demoted by action $a$ respectively as $V_a^+$, and $V_a^-$.

Again, we want to avoid restricting ourselves to particular moral paradigms. We would be doing this if we were to make promotion only signify something like "increasing the quantity of value $v$ after performing action $a$". Nevertheless, note that this might be the meaning used for a utilitarian approach, for instance.

For Hal's action "take insulin", we thus have the set of values which are in line with performing it, as $V_a^+$ = {"Hal's life"} and the set of values which might be demoted by performing it as $V_a^-$ = {"Carla's life", "Property"}.

### 2.3    Impact functions

Based on this, we can now define a function which evaluates the impact that performing a given action $a$ will have on the set of values. We call this the **impact function**. That means we can get a numerical evaluation of how much a given action promotes or demotes a certain value. This will later enable us to compare actions against each other.

Given an action $a$ and a set of values $V$ we define $I_a : V \to \{-1,0,1\}$ as the impact function of action $a$.

$$\mathbf{I}_a(v) = \begin{cases} 1, & if\ v \in V_a^+ \\ -1, & if\ v \in V_a^- \\ 0, & otherwise \end{cases}$$

Impact function of action a.

We call this the impact coefficient of an action with regards to a value. Every action promotes, demotes, or has no impact upon every value. Note that it is possible to extend this function to a larger target set, such as $\mathbb{Z}$ or $[-1,1] \subset \mathbb{R}$, to account for varying degrees of promotion and demotion.

From the above definition, we immediately get the definition of an **impact coefficient** of action $a$ on value $v$:

$$i_a^v := \mathbf{I}_a(v)$$

Impact coefficient of action $a$ on value $v$.

This gives us a convenient way of representing actions in terms of their promotion and demotion of the values in $V$. We represent the effects of actions as tuples, made up of the action's impact coefficients. We call the **impact representation** of an action $a$ the tuple consisting of all the impact coefficients an action has on all the values:

$$\left(i_a^{v_1}, i_a^{v_2}, \dots, i_a^{v_m}\right), \text{where } m = |\boldsymbol{V}|$$

Impact representation of an action $a$.

In Hal's case, using the evaluation of the impact function as above, we would get the representation $a_1$: $(1, -1, -1)$ for action $a_1$. That means that the first action, taking the insulin, would promote the first value, Hal's life, and demote the others.

This allows us to evaluate an action based on the value set given.

## 2.4    Moral Paradigms: Strata with Values

We now have a way of representing actions in terms of their promotion and demotion of relevant values. To make an ethical decision, we also need a way of comparing these actions.

Thus, the second input for MARS is **a moral paradigm** which is an expression of what is considered important and the basis for making ethical decisions. A moral paradigm would tell us, for instance, whether we think being honest is more important than not doing harm or vice versa. It might also tell us whether two values are of comparable importance or whether one value is strictly more important than another one. In MARS, this **qualitative difference between values** is modelled using a stratified ordering of the values.

This qualitative differentiation between values is a central feature of MARS. Using it, one can represent the idea that an action promoting a particular value should always be preferred over an action that promotes lesser values; regardless of the number of lesser values or the amount that they are promoted in. This is crystallized in the concept of **strata** which are layers made up of groups of values that are qualitatively more important than others. For instance, we might have "Life" in a higher stratum than other values and might want any action that saves someone's life to always be preferred over actions that involve other values. A *stratum* is a subset of the value set. The moral paradigm given as input to MARS is in the form of these $k$ strata.

An important advantage of this stratified approach is that we can model situations in which there is no total order over the values or ethical principles in the moral paradigm. Previous work has often operated only under this assumption of having a total order, in approaches such as (Dennis et al. 2016, Bench-Capon 2002).

A stratum $S_i$, $i \in [1, k]$, is a subset of the values in $\boldsymbol{V}$. We define $\boldsymbol{S} = (S_1, S_2, \dots, S_k)$ to be the set of all $k$ strata. This is an input to MARS. $S_i$ is a partition of $\boldsymbol{V}$, so every value in $\boldsymbol{V}$ belongs to exactly one stratum.

We define a relation $>_S$ on strata that helps us enumerate the ordered strata from top to bottom. The higher the stratum, the more important are the values therein. Importantly, note that values being in the same stratum does not necessarily make them equally preferred. This relation has the following properties: Irreflexivity, Asymmetry, Transitivity, Completeness – it is a total order.

Using this relation over strata, we can directly express the qualitative difference between values in different strata. The qualitative difference between values is given by the relation $>_V$. What the formulation of the relation expresses is that a value $v$ is qualitatively more important than value $v'$ if and only if $v$ is in a higher stratum than *v'*. This relation has the following properties: Irreflexivity, Asymmetry, Transitivity, Transitivity of incomparability. It is therefore a strict weak ordering.

How could we model Hal's moral paradigms using strata?

A *selfish version of Hal* might value his own life strictly more than Carla's which could be represented using the following stratified model:

- Stratum 1: Hal's Life
- Stratum 2: Carla's Life
- Stratum 3: Property

An *egalitarian Hal*, however, might value both of their lives similarly which could be modeled using this ordering of values:

- Stratum 1: Hal's Life, Carla's Life
- Stratum 2: Property

### 2.5 Evaluation Models

Given an ordering of values in strata and a set of available actions represented in terms of their impact on those values, let us now consider how MARS selects the ethically preferred one.

This part of MARS, the evaluation algorithm, is the last parameter that can be adjusted to determine the system's functioning.

Values in higher strata are qualitatively more important than those in lower strata, but values within a stratum could still be compared and evaluated against one another, and this is what we describe in this section.

For clarity and brevity, we will illustrate MARS using only two evaluation models: a very basic one, the **Global Maximum Model**, a slightly more advanced one, the **Additive Model**, and a more advanced one yet, the **Weighted Additive Model**, the latter two of which take into account all positive and negative impacts. Afterwards, we briefly touch upon other useful models.

Note that all of these different models yield different desired semantics for the system and can lead to vastly different results. We consider this to be a strong point to MARS, as it is agnostic to the underlying ethical paradigm and can thus flexibly cater to many vastly different ethical standpoints.

**Global Maximum Model.** To begin with, let us model the simplest case, in which we only consider the most important features of each action and in which we have a total order over the values in the moral paradigm. To do this, we can compare actions based on the most important value that the actions impact upon differently. This is what the Global Maximum Model does, and in it the preference relation on actions is defined in the following way: $a_1 \succ_A a_2$ if and only if there is a value for which $a_1$ has a higher impact coefficient than $a_2$ and all values that are higher than it in the moral paradigm are "tied" between the two actions, in that they have the same impact upon them:

$$a_1 \succ_A a_2 \Leftrightarrow \exists v \in \mathcal{V}. i_{a_1}^v > i_{a_2}^v \land \forall v' \in \mathcal{V}. \left( v' \succ_V v \rightarrow i_{a_1}^{v'} = i_{a_2}^{v'} \right)$$

Preference relation on actions in the Global Maximum Model.

When applying this algorithm to selfish Hal's stratified model of values the action $a_1$ "take insulin" would be returned as the preferred option, as its impact coefficient for the value in the topmost stratum ( $i_{a_1}^{v_1}$ ) is 1, while the correspondent ( $i_{a_2}^{v_1}$ ) is $-1$ for action $a_2$.

**Additive Model.** It is apparent that the above model is insufficient to deal with situations where two actions have the same impact coefficient in the highest stratum and does not allow for any complex comparison of actions.

The additive model is based on the intuition that all impact coefficients in a stratum should be considered when evaluating an action. One possible approach to achieving this is to aggregate the impact coefficients, and a simple way of doing that is to aggregate them within strata. To begin, we used addition for this: the sum of impact coefficients of an action for a given stratum is a simple method of evaluating which of the actions is morally preferred, based on a more positive impact upon more of the items that we care about - values. The purpose of this is similar to those that have been proposed before in the context of consequentialism by (Feldman 1997, Dietrich & List 2016).

The preference relation on actions here defines $a \succ_A a'$ if and only if there is some stratum in which the preferred action has more of an impact, measured additively in terms of impact coefficients on the values, than the other actions, and all the corresponding strata before it were "tied":

$$a \succ_A a' \Leftrightarrow$$

$$\exists k \in [1, |\mathcal{S}|]. \sum_{v \in S_k} i_{a'}^v < \sum_{v \in S_k} i_a^v \land$$

$$\forall l \in [1, k-1]. \sum_{v \in S_l} i_{a'}^v = \sum_{v \in S_l} i_a^v$$

Preference relation on actions in the Additive Model.

Let us consider how this model could be applied to find a preferred action according to egalitarian Hal's moral paradigm. As the sum of impact coefficients is the same for both actions in the first stratum, but the action "don't take insulin" has a higher sum in the second stratum, egalitarian Hal will not take the insulin.

Note that this model gives rise to an impact representation of actions evaluated according to the moral paradigm, consisting of this additive aggregation of values within in each stratum. One can then see that an equivalent way of comparing actions is comparing these evaluations using a lexicographic ordering. In the above example "Take insulin" could thus be seen as $(0, -1)$ and "Don't take insulin" could be seen as $(0,1)$. This representation makes it easier to see why the second action will in this case be preferred.

**Weighted Additive Model.** The additive model can be extended by weighing impact coefficients differently. This would model a moral paradigm in which two values in the same stratum are comparable, but not equally important, giving us a way of quantitatively differentiating them. Furthermore, Dancy (Dietrich & List 2016) suggests that the importance of features that are morally relevant might change depending on the situation. Thus, we can use this concept of weights to elegantly model this changing importance of the different values.

We define weights $w^v \in R, v \in \mathcal{V}$, associated with the impact coefficients $i_a^v, \forall a \in A$ and the relation $\succ_A$ for the additive model as:

$$a \succ_A a' \Leftrightarrow$$

$$\exists k \in [1, |\mathcal{S}|]. \sum_{v \in S_k} w^v i_{a'}^v < \sum_{v \in S_k} w^v i_a^v \wedge$$

$$\forall l \in [1, k-1]. \sum_{v \in S_l} w^v i_{a'}^v = \sum_{v \in S_l} w^v i_a^v$$

Preference relation on actions in the Weighted Additive Model.

**Preferred Actions Set.** So far, we have shown how we can do pairwise comparison of actions. We also have semantics for a **preferred actions** set $P$, being the set of actions which are preferred over all available actions, which sets out that an action $a \in P$ if and only if there is no other action which is preferred to it.

$$a \in \mathcal{P} \Leftrightarrow \nexists a' \in A. a' \succ_A a$$

Condition for inclusion of action $a$ in the set of preferred actions $P$.

Note that there might be several actions in this set.

**Alternative Models.** For brevity, we shall only mention some of the other evaluation models we have for MARS.

*Minimal Number of Negative Impact Coefficients:* We might want to prefer actions which demote as few values as possible, where violations in higher strata are deemed more important than violations in lower strata.

*Minimal Sum of Demotions*: Similarly, we might want to prefer actions with minimal sums of demotion.

*Stratum Satisfaction and Violation:* What about scenarios in which there are several ways of achieving a goal or of promoting a value? The core idea is that if an available action promotes one of the values in a stratum, that entire stratum is satisfied, and the other values are no longer considered.

**Enjoying MARS: Workflow.** When using MARS, we use the following approach:

1. Establish a scenario to model. Establish what the available actions $A$ are.
2. Decide which values or features the decision depends on. Such features could be intentions, reasons, desirable consequences etc. These values make up the set $V$.
3. Establish which values are of comparable importance and which are qualitatively more important than others. This will yield the ordering of values in strata, $S$.
4. Decide how the actions impact upon the values. Is performing a given action "in the spirit of" a value? Does the action affect the value positively? If so, the impact coefficient for that pair should be 1. If it affects it negatively, or acts "against the spirit" of the value, then it should be a -1. If the value is irrelevant to the action it is a 0. Following this approach, we obtain the impact representation of the actions, given by the impact functions in $I$.

5. Select an evaluation model, to compare the actions against each other. Do we want values in the same stratum to be counted or added up, or weighted and compared against each other? Do we want to select actions that have the most good impact coefficients, or the least negative impact coefficients?

6. Apply the evaluation model to obtain the set of preferred actions $P$. These are the actions one can perform, given the stratified moral paradigm used as input.

## 3    Related work

Firstly, the *GenEth* system by Anderson et al. uses a similar action representation to ours (Anderson & Anderson 2014). In their work, a bottom-up approach of learning ethical principles is presented. We, on the other hand, introduce a top-down approach. As such, the systems offer the respective advantages and disadvantages of these strategies, and we agree with the previous assessment that a combination of top-down and bottom-up is necessary to create moral artificial agents (Charisi et al. 2017). Top-down systems allow for an a priori, explicit encoding of principles which allows for an informed decision-making process and possibly verification of the system before it is employed. On the other hand, we believe that encoding moral paradigms and making the connection between abstract principles and sensory inputs might prove hard, posing a challenge for such a top-down approach.

Dennis et al. (Dennis et al. 2016) propose a hybrid system implementing both a component to interact with the environment to obtain knowledge about the actions, as well as a discrete decision-making component. However, the ethical policy provided to the agent in their system consists of a total order over the ethical principles. Our system, in contrast, does not require a total order of values, thanks to the stratified structure of the moral paradigm, thus allowing us to encode more varied moral paradigms more easily. This is useful when there is uncertainty about the way that the values ought to be ranked. We can still cater for the special case of a total order on values and get similar results to the authors.

Pereira et al. (Pereira et al. 2016) tackle automated ethical reasoning by demonstrating how logic programming is a suitable paradigm to model several relevant aspects in moral reasoning. We see our work as complementary to theirs, in that we do not directly address the same problem in the same way, and we do not focus on the underlying implementation and its properties but on the higher-level theory and model used for reasoning. Furthermore, they present a more sophisticated and nuanced approach with respect to moral permissibility which we only touch upon.

We have tested MARS using the above related work, attempting to faithfully recover the inputs used therein and apply them to our system, checking to see if we can obtain the same results in terms of preferred actions. We have found that MARS is flexible enough to allow for the different semantics found therein and, by selecting a suitable evaluation model, can be tweaked to recreate the results obtained in the works above. Note that, as mentioned above, we see our work as complementary to their approaches, as the aim of our system is not identical to any of theirs.